%% file: main.tex
\documentclass[conference]{IEEEtran}
\usepackage{cite}
\usepackage{amsmath,amssymb,amsfonts}
\usepackage{algorithmic}
\usepackage{listings}
\usepackage[hidelinks]{hyperref}
\usepackage{graphicx}
\usepackage{textcomp}
\usepackage{xcolor}
\usepackage{tabularx}
\usepackage{booktabs}
\usepackage{tikz}
\usepackage{makecell}
\usepackage{svg}
\usepackage{caption}
\usepackage{subcaption}
\usepackage{lipsum}
\usepackage{comment}
\usepackage[numbers]{natbib}

\usepackage[nolist]{acronym}
\usepackage{float}
\usepackage[ruled,vlined]{algorithm2e}

\lstdefinestyle{xmlstyle}{
    basicstyle=\ttfamily\scriptsize,  % Schriftgröße und Typewriter-Stil
    breaklines=true,                    % Zeilenumbrüche erlauben
    frame=none,                        % Rahmen um den Code weg
    captionpos=b,                        % Caption unterhalb
    language=XML,                    % XML-Syntaxhervorhebung
    aboveskip=0pt,
    belowskip=0pt
}
\lstdefinelanguage{Turtle}{
    morekeywords={@prefix,@base,a,rr,rml,ql,math,xsd},
    keywordstyle=\color{blue},    % Blau für Schlüsselwörter
    sensitive=true,
    morecomment=[l]{\#},           % Kommentare beginnen mit #
    morestring=[b]"",
}
\lstdefinestyle{rmlstyle}{
    basicstyle=\ttfamily\scriptsize,   % Kompakte Schriftgröße
    breaklines=true,                    % Automatische Zeilenumbrüche
    frame=none,                          % Kein Rahmen
    backgroundcolor=\color{gray!10},     % Leicht grauer Hintergrund
    captionpos=b,                        % Caption unterhalb
    language=Turtle,                     % Turtle für RML-Notation
    emph={xsd:double, xsd:integer},      % Datentypen fett hervorheben
    emphstyle=\bfseries
}

\def\BibTeX{{\rm B\kern-.05em{\sc i\kern-.025em b}\kern-.08em
    T\kern-.1667em\lower.7ex\hbox{E}\kern-.125emX}}

\definecolor{light-gray}{gray}{0.95}
\lstdefinestyle{mystyle}{
    backgroundcolor=\color{light-gray},  
    basicstyle=\ttfamily \footnotesize,
    breaklines=true
}

\begin{document}
\bstctlcite{IEEEexample:BSTcontrol}
\begin{acronym}

  \acro{llm}[LLM]{\textit{Large Language Model}}
  \acroplural{llm}[LLMs]{\textit{Large Language Models}}

  \acro{odp}[ODP]{\textit{Ontology Design Pattern}}
  \acroplural{odp}[ODPs]{\textit{Ontology Design Patterns}}
  
  \acro{nlp}[NLP]{\textit{Natural Language Processing}}

  \acro{nl}[NL]{\textit{Natural Language}}

  \acro{rnn}[RNN]{\textit{Recurrent Neural Network}}
  \acroplural{rnn}[RNNs]{\textit{Recurrent Neural Networks}}
  
  \acro{sparql}[SPARQL] {\textit{SPARQL Protocol and RDF Query Language}}

  \acro{http}[HTTP]{\textit{Hypertext Transfer Protocol}}

  \acro{kg}[KG]{\textit{Knowledge Graph}} 
  \acroplural{kg}[KGs]{\textit{Knowledge Graphs}}
  
  \acro{owl}[OWL]{\textit{Web Ontology Language}}

  \acro{cps}[CPS]{\textit{Cyber-Physical System}}
  \acroplural{cps}[CPS]{\textit{Cyber-Physical Systems}}

  \acro{cq}[CQ]{\textit{Competency Question}}
  \acroplural{cq}[CQs]{\textit{Competency Questions}}

  \acro{uri}[URI]{\textit{Uniform Resource Identifier}}
  \acroplural{uri}[URIs]{\textit{Uniform Resource Identifiers}}

  \acro{scq}[SCQ]{\textit{standard-compliant question}}
  \acroplural{scq}[SCQs]{\textit{standard-compliant questions}}

  \acro{nscq}[NSCQ]{\textit{non-standard-compliant question}}
  \acroplural{nscq}[NSCQs]{\textit{non-standard-compliant questions}}
\end{acronym}

\title{Representing Time-Continuous Behavior of Cyber-Physical Systems in Knowledge Graphs}

\author{
    \IEEEauthorblockN{
        Milapji Singh Gill\IEEEauthorrefmark{1},
        Tom Jeleniewski\IEEEauthorrefmark{1},
        Felix Gehlhoff\IEEEauthorrefmark{1},
        Alexander Fay\IEEEauthorrefmark{3}%
    } 
    
    \IEEEauthorblockA{
        \IEEEauthorrefmark{1}Institute of Automation Technology\\
        \textit{Helmut Schmidt University Hamburg, Germany}\\
        {\small \{milapji.gill, tom.jeleniewski, felix.gehlhoff\}@hsu-hh.de}\\
        \IEEEauthorrefmark{3}Chair of Automation\\
        \textit{Ruhr University, Bochum, Germany}\\
        {\small alexander.fay@rub.de}\\
       }
       \vspace{-1cm}
}

\maketitle

\begin{abstract}
Time-continuous dynamic models are essential for various \ac{cps} applications. To ensure effective usability in different lifecycle phases, such behavioral information in the form of differential equations must be contextualized and integrated with further CPS information. While knowledge graphs provide a formal description and structuring mechanism for this task, there is a lack of reusable ontological artifacts and methods to reduce manual instantiation effort. Hence, this contribution introduces two artifacts: Firstly, a modular semantic model based on standards is introduced to represent differential equations directly within knowledge graphs and to enrich them semantically. Secondly, a method for efficient knowledge graph generation is presented. A validation of these artifacts was conducted in the domain of aviation maintenance. Results show that differential equations of a complex Electro-Hydraulic Servoactuator can be formally represented in a knowledge graph and be contextualized with other lifecycle data, proving the artifacts' practical applicability.
\end{abstract}


\input{Text/01_Introduction}

\input{Text/02_Background}

\input{Text/03_Related_Work}

\input{Text/04_Modeling_Behavioral_Models}

\input{Text/05_Application}

\input{Text/06_Discussion}

\input{Text/07_Summary_and_Outlook}

\section*{Acknowledgment}
This research [project ProMoDi and LaiLa] is funded by dtec.bw – Digitalization and Technology Research Center of the Bundeswehr. dtec.bw is funded by the European Union – NextGenerationEU.

\input{bibliography/References.bbl}

\end{document}

%% file: Text/01_Introduction.tex
\section{Introduction}
\label{sec:introduction}

\acp{cps} are a key enabler of Industry 4.0, forming the backbone of intelligent, interconnected, and automated industrial processes. These systems tightly integrate physical components with digital control, enabling seamless interaction between hardware and software \cite{VDIVDE2206}. CPS applications span the entire system lifecycle, requiring various information about the system \cite{Sabou.2020}. One specifically important information class for several use cases are behavioral models, capturing system dynamics \cite{Derler.2012}. In the early phases, simulations are used to analyze the behavior of CPS, aiding in design, virtual commissioning, and optimization \cite{Stegmaier.2022}. During operation, a combination of system dynamics and data-driven approaches is essential for tasks such as anomaly detection and fault diagnostics \cite{Westermann.2023, Gill.2023a}. Consequently, a unified and formalized representation of system behavior integrated with further lifecycle data not only enhances knowledge management but enables engineers to use this knowledge to predict, analyze, and optimize CPS. 

In this context, ontologies form a robust foundation for constructing knowledge graphs and realizing the required formal representation of CPS \cite{Hildebrandt.2020}. By providing well-defined semantics and systematically organizing domain knowledge, ontologies facilitate seamless data integration, interoperability, and the reuse of knowledge \cite{InternationalElectrotechnicalCommission.07.04.2025}. These capabilities are valuable in all CPS lifecycle phases including engineering, operation and maintenance, where data is often fragmented across heterogeneous sources \cite{Sabou.2020}. 
However, formalizing and contextualizing CPS behavior with knowledge graphs presents several hurdles. One of them is the diversity of modeling approaches used to describe CPS behavior. While there are many approaches to formalize discrete state models \cite{Westermann.2023, Bakirtzis.2022, Hildebrandt.2017}, there is a lack of approaches to include differential equations, which are necessary for modeling continuous system dynamics \cite{Derler.2012}. Furthermore, it requires extensive effort to develop such models.  Ontologies are inherently complex and not easily understandable, making them difficult for end users to engage with effectively. This complexity becomes even more problematic in light of the shortage of ontology experts \cite{Hildebrandt.2020}. Thus, this work investigates the following research questions (RQ):

\begin{itemize}
\item \textbf{RQ1:} \textit{How can differential equations be represented in knowledge graphs to describe complex continuous dynamic behavior and be contextualized with further relevant CPS information?}
\item \textbf{RQ2:} \textit{How can modeling activities be partially automated to reduce the manual effort required for developing knowledge graphs that represent continuous behavior?}
\end{itemize}
\vspace{-0.01 cm}

The remainder of the paper is structured as follows: Sec.~\ref{sec:Background} provides background about CPS behavioral models as well as ontologies in order to derive relevant requirements. Sec.~\ref{sec:RelatedWork} presents an analysis of related work, identifying gaps in the current state of research. Sec.~\ref{sec:Modeling} introduces the proposed artifacts to answer the RQs posed before, detailing their design and function. Sec.~\ref{sec:application} demonstrates the applicability of the approach through a case study in the aviation domain, focusing on the maintenance of aircraft components. Sec.~\ref{sec:discussion} discusses the results and implications of the findings. Finally, Sec.~\ref{sec:conclusion} concludes with a summary and an outlook on future work.

%% file: Text/02_Background.tex
\section{Background and Requirements}  \label{sec:Background}

\subsection{Behavioral Models of Cyber-Physical Systems} \label{sec:Behavior}

The behavior of CPS can be described using various modeling approaches depending on the nature of the system, its control requirements, and the viewpoint regarding the process granularity. Generally, the modeling depth ranges from time and space discrete to time and space continuous, each suited for specific aspects of CPS functionality \cite{Stegmaier.2022}. 

For CPS, primarily governed by control logic, discrete behavior, depicted for example with State Machines or Petri Nets, is widely used, when temporal effects are not considered in detail. State Machines describe system behavior through well-defined states and transitions triggered by events. Petri Nets extend this concept by incorporating concurrent events and synchronization mechanisms, enabling the modeling of complex, event-driven CPS. Their expressive power allows for the representation of parallelism, resource allocation, and deadlock detection, making them particularly effective for formal verification and logical consistency analysis \cite{Wisniewski.2021}. When temporal aspects become relevant, timed extensions (see Timed Automata) can be incorporated into states, transitions, or tokens to capture time-dependent behavior \cite{Westermann.2023}.

In contrast, continuous physical processes (e.g., mechanical, fluid, electromechanical) necessitate different modeling approaches than those used for discrete behavior \cite{Derler.2012}. To formally model time-continuous behavior, mathematical formulations such as differential equations are commonly employed. Ordinary differential equations (ODEs)  express temporal aspects in system dynamics, depending only on a single independent variable \cite{Stegmaier.2022, Taha.}. Partial differential equations (PDEs) extend this concept to systems where state variables depend on multiple dimensions, capturing more complex physical interactions \cite{Taha.}.

For improved CPS understanding, it is advantageous to integrate behavioral information with CPS functions and structure. In this context, behavior, including dynamic properties, procedures, states, and variable interactions, is mapped to specific functions and structural elements. Functions define the system's intended process in a solution-neutral manner by transforming inputs (signals, products, or energies) into outputs, while the structure comprises the physical or logical components and their interconnections \cite{Hildebrandt.2017}.

\subsection{Domain-Specific Ontologies} \label{sec:Ontologies}

Ontologies provide a unified semantic structure for integrating and linking heterogeneous data sources. Beyond integration, ontologies enable the formalization of domain-specific knowledge, capturing expert insights in a machine-readable format. This structured representation ensures that knowledge is preserved, reusable, and interpretable by both humans and automated systems \cite{Hildebrandt.2020}. Furthermore, ontologies provide unambiguous semantics that enhance interoperability between heterogeneous systems, making them particularly valuable in complex industrial environments where multiple tools and platforms must exchange information reliably. By defining formal relations between concepts, ontologies enable automated inference to detect inconsistencies, infer missing information, and validate system configurations \cite{InternationalElectrotechnicalCommission.07.04.2025}. 

CPS generate and utilize information from various domains, including systems engineering, real-time operational data, and maintenance records \cite{Gill.2024}. Thus, domain-specific ontologies offer a semantically rich approach to modeling CPS and linking information across different lifecycle phases \cite{Sabou.2020, Gill.2024}. To minimize modeling effort and maximize reusability, a modular approach to ontology design is essential. In this context Ontology Design Patterns (ODPs) are essential, which are based on established standards with a unified vocabulary of concepts \cite{Hildebrandt.2020}. This modularity supports efficient ontology evolution, enabling application-specific extensions while maintaining compatibility with existing domain-specific models. 

\subsection{Requirements for Formalizing CPS Behavioral Models} \label{req}

Building on the insights from the previous sections, the following requirements (R) have been derived.

\noindent \textbf{(R1) Representation of differential equations:}
Differential equations are fundamental for modeling complex physical behavior in CPS \cite{Taha.}. Thus, a semantic model must be capable of describing time-continuous system dynamics such as ODEs and PDEs using mathematical operators.  

\noindent \textbf{(R2) Contextualization with further information from the CPS lifecycle:}
The formalized behavioral knowledge must support the association with further specific system information like function and structure which are gathered throughout the CPS lifecycle \cite{Hildebrandt.2017, Derler.2012}. This alignment enables a modular system model where the behavior is put into context in order to improve the system understanding.

%Mächtigkeit der FPB
\noindent \textbf{(R3) Composable and flexible process models at different granularity levels:}
To effectively model the behavior of CPS within knowledge graphs, it is essential to be able to capture different modeling depths of behavior \cite{Stegmaier.2022}. This requires modular and hierarchical process models, allowing both composition and decomposition of sub-processes to represent different levels of process granularity \cite{Jeleniewski.2024}.

\noindent \textbf{(R4) Semantic formalization with structured serialization:}
The knowledge should be encoded in a formal semantic model that adheres to a usable serialization format, such as XML, JSON-LD, or RDF. Such machine-readable formats enable interoperability, automation, and facilitate data exchange across various tools and platforms \cite{InternationalElectrotechnicalCommission.07.04.2025}.

\noindent \textbf{(R5) Standards-based description of concepts:}
The semantic model should leverage existing standards. A standards-based description of relevant concepts and relations fosters interoperability between different tools, systems, and domains across the CPS lifecycle \cite{Hildebrandt.2020}. 

\noindent \textbf{(R6) Modular and extensible model:} The semantic model must support modularity and extensibility, allowing incremental updates and additions depending on the application scenario. A modular approach, e.g. with ODPs, ensures that the model can accommodate changes without requiring a complete overhaul of the existing model \cite{Hildebrandt.2020}.

%% file: Text/03_Related_Work.tex
\section{Related Work} \label{sec:RelatedWork}
This section analyzes related work with respect to the requirements (R1–R6) from Sec.~\ref{sec:Background}, identifying how existing approaches address or lack support for these aspects.

\citet{Cheong.2019} propose a physics‑based simulation ontology to capture the key elements required to model simulation input data for physical systems. While this ontology clearly defines domains, material properties, and boundary conditions for constructing differential equations, it does not provide a semantic representation for the mathematical operators (e.g. derivatives). Moreover, the paper primarily focuses on the simulation aspect of CPS. While it briefly touches upon representing functions at different process granularities (R3), behavior contextualization within the overall CPS is only partially considered. R4-R6 are addressed through a standards‑based, modular, and extensible ontology with usable serialization formats.

In the domain of semantic modeling for production systems, \citet{Hildebrandt.2017} have made significant contributions by structuring systems based on function, structure, and behavior (R2, R3). Moreover, a method was introduced for creating and aligning modular ODPs, built on industrial standards (R4, R5, R6)  that semantically depict knowledge in the production domain \cite{Hildebrandt.2020}.  Notably, the behavioral models in this methodology are represented as discrete event models, capturing the system's behavior in terms of distinct state transitions. Differential equations, in contrast, are not considered. 

\citet{Bakirtzis.2022} propose an ontological metamodel (R3) that captures system behavior using discrete control models and event‐based failure scenarios, linking loss scenarios and unsafe control actions to resilience mechanisms such as anomaly detection. Moreover, the behavior is contextualized with structural and functional information (R2). The approach builds on established industry metamodels and standards (R5) and is designed with a modular and extensible semantic model (R6). However, the representation of system behavior at varying process granularities is only marginally addressed. Additionally, it does not explicitly incorporate continuous dynamics through ODEs or PDEs. 

\citet{Kovalenko.} present the AutomationML Ontology, a semantic representation of the AutomationML standard for Cyber-Physical Production Systems (R4, R5, R6). The ontology enhances engineering data integration, consistency checking, and hierarchical reasoning, supporting interoperability. System behavior is depicted using discrete representations, focusing on functional, structural, and interface relations (R2, R3), but does not depict differential equations (R1). 

\citet{Westermann.2023}, building on a method introduced by \citet{Gill.2023a}, combine data-driven learning with semantic modeling by extracting timed automata from operational CPS data and linking them to ontologies. Their methodology supports contextualization of discrete behavior with functional and structural information for anomaly detection (R2, R3) and applies ODPs grounded in industrial standards (R4, R5, R6). However, time-continuous dynamics such as ODEs or PDEs is not incorporated (R1).

In summary, while nearly all of the presented approaches use modular, standards-based ontologies for formalization (R4–R6) and incorporate additional CPS context (R2–R3), the focus remains predominantly on discrete behavioral models. The semantic enrichment of ODEs and PDEs (R1) is only marginally addressed. Only \citet{Cheong.2019} include differential equations, and even there, the specification of the necessary mathematical operators is omitted. In conclusion, this research gap needs to be addressed.

%% file: Text/04_Modeling_Behavioral_Models.tex
\section{Representing Time-Continuous Behavior of CPS in Knowledge Graphs} \label{sec:Modeling}

In the following, two artifacts are presented to address the RQs posed. First, a semantic model is introduced to represent and contextualize time-continuous dynamics in knowledge graphs (RQ1). Second, a mapping method for partially automating the instantiation of the behavioral model is proposed (RQ2).

\subsection{Semantic Model for CPS Behavior Contextualization}

\begin{figure*}
    \centering
    \includegraphics[width=0.9\textwidth]{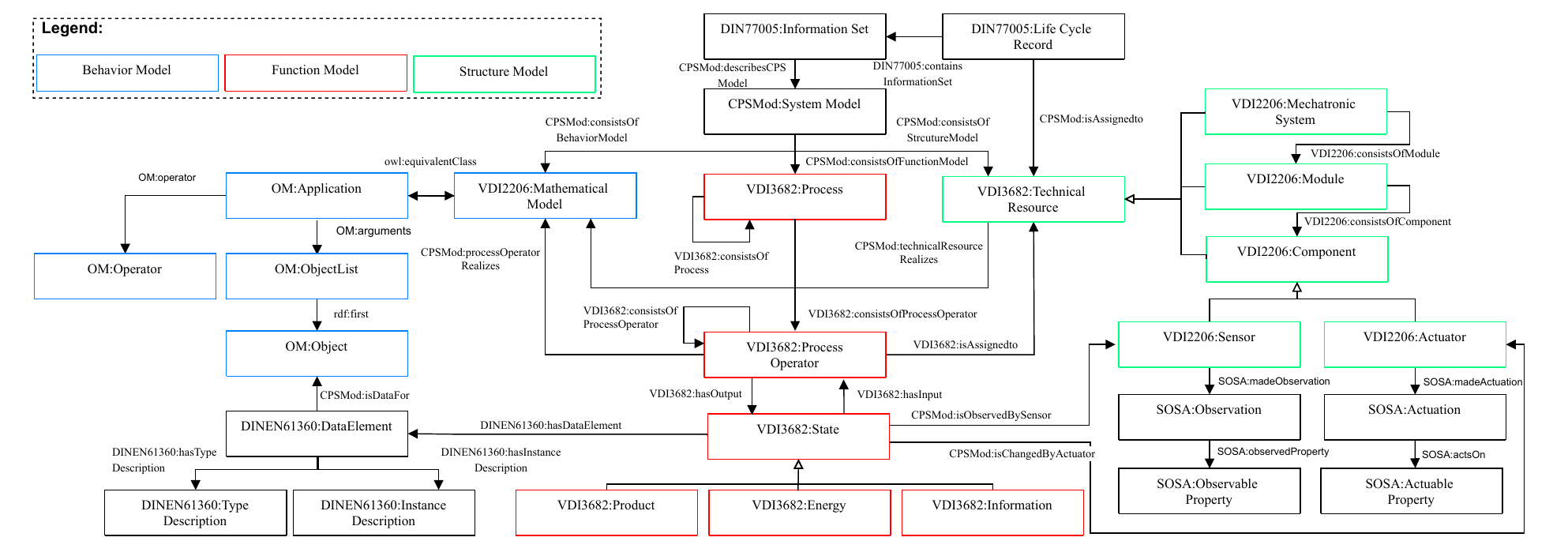}
    \caption{Semantic model for CPS behavior contextualization}
    \label{fig:SemanticModel}
\end{figure*}

 The semantic model \texttt{CPSMod} (see Fig.\ref{fig:SemanticModel}) is an alignment ontology composed of various ODPs. Most of these ODPs were developed in earlier works targeting different application domains of CPS and are now being reused \cite{Gill.2023, Jeleniewski.2023, Hildebrandt.2020}. A curated set of these ODPs can be found on \textit{GitHub}\footnote{\url{https://github.com/hsu-aut/Industrial-Standard-Ontology-Design-Patterns}}. First, a lifecycle record is created for each CPS under consideration according to the ODP DIN 77005. Using the information classes \texttt{DIN77005:LifeCycleRecord} and \texttt{DIN77005:InformationSet}, CPS models and data can be structured and collected across different lifecycle phases, such as engineering, operation, and maintenance. In the considered application case, the \texttt{CPSMod:SystemModel} is integrated into the lifecycle record as a \texttt{DIN77005:InformationSet}. It encompasses the three primary information classes of systems engineering: function (depicted in red), structure (depicted in green), and behavior (depicted in blue) \cite{Hildebrandt.2017}. In order to achieve a modular and reusable structure, the semantic model aligns ODP VDI~3682 for functional modeling, ODP VDI 2206 for structural representation, and ODP DIN EN 61360 for specifying CPS variables and parameters in mathematical modeling. Additionally, it uses the OpenMath standard as an ODP to semantically describe these mathematical models \cite{Wenzel.}, particularly differential equations. Additionally, the SOSA (Sensor, Observation, Sample, and Actuator) ontology is incorporated to depict the actual sensor and actuator states of the operating CPS. 

The ODP VDI 3682 provides a basis to model the relevant  \texttt{VDI3682:Process}. The class \texttt{VDI3682:ProcessOperator} represents separate functions within the \texttt{VDI3682:Process}. The ODP enables the composition and decomposition of \texttt{VDI3682:Process} and \texttt{VDI3682:ProcessOperator}. This allows for the representation of individual functionalities including assigned differential equations at different process granularities, depending on the requirements of the application case. Additionally, parallel and alternative branches can be modeled within \texttt{VDI3682:Process}, enabling a flexible and structured depiction of process flows. To ensure correct functional dependencies, the connection between separate \texttt{VDI3682:ProcessOperator} is established through the input and output relations of \texttt{VDI3682:State}, specifically the flow of \texttt{VDI3682:Product}, \texttt{VDI3682:Energy}, or \texttt{VDI3682:Information}. The alignment of ODPs VDI 3682 and IEC 61360  with the object property \texttt{DINEN61360:hasDataElement} provides a standardized schema for defining and precising a \texttt{DINEN61360:DataElement} in relation to \texttt{VDI3682:State} as well as the \texttt{VDI3682:TechnicalRessource}. This allows to specify type descriptions (\texttt{DINEN61360:TypeDescription}) that apply to data elements as well as instance descriptions that specify properties assigned to individual data elements (\texttt{DINEN61360:InstanceDescription}). The \texttt{DINEN61360:DataElement} class is essential for mathematical modeling, as it enriches equations semantically with regard to relevant input/output variables and parameters. 

Each \texttt{VDI3682:ProcessOperator} is additionally linked to a \texttt{VDI3682:TechnicalResource}, enabling the assignment of CPS functions to specific structural components. This alignment ensures that the execution of functional requirements is directly associated with the physical or logical elements of the CPS. Each \texttt{VDI3682:TechnicalResource} can be specified using ODP VDI 2206. A \texttt{VDI3682:TechnicalResource} assigned to a \texttt{VDI3682:ProcessOperator} can represent the entire \texttt{VDI2206:MechatronicSystem}, but it can also refer to more detailed structural elements, such as an individual \texttt{VDI2206:Module}, or even \texttt{VDI2206:Component}. This enables the semantically enriched representation of mathematical models at different levels of system structure, providing a comprehensive view of the CPS dynamic interactions. 

To finally capture the dynamic interactions within a CPS, OpenMath is utilized as indicated. For this purpose, the \texttt{VDI3682:ProcessOperator} is linked to the \texttt{VDI2206:MathematicalModel} class, which, according to the standard, serves as the "\textit{[...] basis for describing the behavior of systems}" \cite{VDIVDE2206}. This primarily refers to physical models, which transform the black-box representation of a \texttt{VDI3682:ProcessOperator} into a white-box model by explicitly defining the relation between inputs and outputs through the \texttt{OM:Application} class. Each functional transformation, as described in VDI 3682, is associated with a mathematical representation of its behavior through \texttt{OM:operator}, which links variables and parameters using \texttt{OM:ObjectList} and \texttt{OM:arguments}. By leveraging the \textit{Content Dictionaries} (CDs) \texttt{weylalgebra1}, more complex equations such as the required ODEs and PDEs can also be included in the knowledge graph. This library is particularly suitable for modeling a wide range of differential equations arising in various processes, including mechanical dynamics, thermodynamic systems, and control systems. A list of the relevant operators, including those essential for behavioral models, is available at the following URL\footnote{\url{https://openmath.org/cd/weylalgebra1.html}}, for example:

\begin{itemize}
  \item \texttt{diff} – ordinary differentiation of a function in one variable (e.g., $\frac{d x^2}{dx} = 2x$)
  \item \texttt{partialdiff} – partial differentiation of multivariable functions (e.g., $\frac{\partial^2 x^2 y}{\partial x^2} = 2y$)
\end{itemize}

While \texttt{weylalgebra1} provides core operators for modeling time-continuous behavior, more complex models may require additional OpenMath CDs. \texttt{calculus1} includes integration operators (e.g. \texttt{int}) essential for cumulative effects, while \texttt{transc1} defines nonlinear functions (e.g., \texttt{sin}, \texttt{exp}) commonly used in physical models. For stochastic behavior, \texttt{stats1} offers constructs for probability and uncertainty. These CDs enable a richer, semantically precise representation of hybrid, nonlinear, and uncertain system dynamics in knowledge-based CPS models. 
In addition to the OpenMath RDF mathematical relations, real system behavior is integrated using the SOSA ontology, which standardizes sensor observations and actuator processes. By leveraging \texttt{SOSA:Observation} and \texttt{SOSA:Actuation} with timestamps, measured system dynamics are semantically linked to the modeled equations.

In conclusion, the presented approach fulfills all defined requirements (see Sec. \ref{req}). A modular, standards-based semantic model \texttt{CPSMod} was introduced as an alignment ontology (see R4 - R6). It integrates OpenMath for representing differential equations within knowledge graphs (R1) and applies established ODPs, such as VDI 3682 and VDI 2206, to contextualize modeled behavior (R2). Especially ODP VDI 3682 and VDI 2206 support the composition of flexible process models across different levels of granularity, depending on the CPS structure (R3).

\subsection{Method for the Instantiation of the Behavioral Model}
To systematically and efficiently instantiate and integrate the depicted behavioral model within the presented semantic model, we extend the method introduced by \citet{Jeleniewski.2023}, which adheres to systems engineering principles. The CPS process model is incrementally constructed according to \citet{Jeleniewski.SemanticModel2023} including functions and connecting them to structure first, as well as secondly modeling the behavior. This method establishes mathematical relations between the inputs and outputs of a \texttt{VDI3682:ProcessOperator} using OpenMath, based on the formalized process description defined in VDI 3682. By initially defining and allocating the class \texttt{VDI3682:ProcessOperator} within the CPS structure (using \texttt{VDI:3682:TechnicalResource}), the method progressively develops the process model at the required level of abstraction, integrating key flows of energies, information, and products. The instances of the class \texttt{VDI3682:ProcessOperator} are then analyzed, decomposed if necessary, and connected to form a consistent process model across different system hierarchies.
For further details on the individual steps of the method, please refer to the original paper, where they are explained in detail \cite{Jeleniewski.2024}. 

To support the inclusion of differential equations in the process model, further steps are taken to incorporate an instantiation method for the mathematical model. After defining and connecting all relevant \texttt{VDI3682:ProcessOperator} elements along with their respective inputs and outputs at the required level of process granularity, the mathematical modeling phase begins. For each \texttt{VDI3682:ProcessOperator}, the engineer formulates differential equations that describe the CPS behavior, based on the associated inputs, outputs, and parameters of the corresponding \texttt{VDI3682:TechnicalResource}.
Once all equations have been defined, a tool-assisted approach is employed to instantiate the behavioral model. The equations are first converted into an OpenMath XML representation, which encodes the complete mathematical structure in a machine-readable format. For this task, already existing tools can be used to automatically generate an OpenMath XML\footnote{\url{https://openmath.org/software/}} file. Mathematical expressions encoded in OpenMath XML exhibit a highly structured, recursive tree architecture. Each expression is composed of applications (\texttt{<OMA>}), symbols from standardized content dictionaries (\texttt{<OMS>}), and variables (\texttt{<OMV>}). Argument sequences are strictly ordered and nested, enabling the representation of complex constructs such as derivatives, algebraic operations, or compositions. Standard RDF mapping mechanisms, such as the RDF Mapping Language (RML) introduced by \citet{Dimou2014}, are not well suited to this type of structure. These tools lack native capabilities for structural recursion, making it difficult to traverse and interpret arbitrarily nested application trees. 

To address these limitations, we apply a recursive tree traversal algorithm\footnote{\url{https://github.com/MilapjiSinghGill/OpenMathXML-RDF-pymapper}} that transforms OpenMath XML into RDF. The algorithm identifies node types dynamically and constructs an RDF graph while preserving syntactic structure and semantic roles. The transformation is as follows: Each function application is represented as an \texttt{om:Application} node. The first child of each application is treated as the operator and linked via \texttt{om:operator}. All remaining children are interpreted as arguments and serialized into an RDF list structure using \texttt{rdf:first}, \texttt{rdf:rest}, and \texttt{rdf:nil}, referenced via \texttt{om:arguments}. Variables and constants are mapped to appropriate RDF nodes with semantic typing (e.g., \texttt{om:Variable}, \texttt{om:Literal}). These core operations are realized by two key functions (see Algorithm \ref{alg:OM2RDF}): \texttt{processnode()}, which recursively classifies and transforms each XML element based on its OpenMath type, and \texttt{createRDFlist()}, which assembles ordered argument sequences into valid RDF list structures. 

\begin{algorithm}[!t]
\caption{Recursive transformation of OpenMath XML to RDF}
\label{alg:OM2RDF}
\scriptsize
\KwIn{OpenMath XML node \texttt{node}}
\KwOut{Corresponding RDF node}

\SetKwFunction{FProcess}{processnode}
\SetKwFunction{FList}{createRDFlist}

\SetKwProg{Fn}{Function}{:}{}
\Fn{\FProcess{node}}{
    \uIf{node is an application (OMA)}{
        operator $\leftarrow$ \FProcess{first child}\;
        arguments $\leftarrow$ list of \FProcess{each remaining child}\;
        rdfList $\leftarrow$ \FList{arguments}\;
        \Return{applicationnode(operator, rdfList)}
    }
    \uElseIf{node is a symbol (OMS)}{
        \Return{URI from content dictionary and name}
    }
    \uElseIf{node is a variable (OMV)}{
        \Return{RDF node of type \texttt{om:Variable} with name}
    }
    \uElseIf{node is a literal (OMI or OMF)}{
        \Return{RDF node of type \texttt{om:Literal} with value}
    }
}

\vspace{0.5em}
\Fn{\FList{elements}}{
    \uIf{elements is empty}{
        \Return{\texttt{rdf:nil}}
    }
    \Else{
        head $\leftarrow$ RDF node with \texttt{rdf:first = elements[0]}, \\
        \phantom{head $\leftarrow$ } \texttt{rdf:rest =} \FList{elements[1:]}\;
        \Return{head}
    }
}
\end{algorithm}

Subsequently, the \texttt{VDI3682:ProcessOperator} is linked to the RDF-serialized differential equations via the object property \texttt{CPSMod:processOperatorBehaviorModel}. Additionally, each variable or parameter within the \texttt{OM:Application} is assigned to a previously defined \texttt{DINEN61360:DataElement} through the object property \texttt{CPSMod:isDataFor}. The result is a process model of the CPS, in which the previously black-box functions of the CPS are refined through their corresponding RDF-serialized behavioral model.

%% file: Text/05_Application.tex
\section{Exemplary Application} 
\label{sec:application}

Understanding the function and behavior of critical components is key to efficient fault diagnostics in aviation maintenance \cite{Gill.2023}. Electro-Hydraulic Servo Actuators (EHSA) in primary flight control systems, for example, can significantly affect aircraft reliability, safety, and maintenance costs when faults occur \cite{Ritter.2018}. Incomplete documentation and fragmented information flows among manufacturers, operators, and maintenance teams often lead to ineffective diagnostics due to missing insights on system dynamics and potential faults. Formalizing engineering knowledge in a queryable knowledge graph enables simulation of system behavior, assessment of degradation, and data-driven diagnostics. The following knowledge graph builds on the previous work by \citet{Ritter.2018}, who modeled the behavior of the EHSA and its components in detail to generate test signals and key features for fault diagnostics.  Adding semantics to these models enables reuse as well as integration with operational data and diagnostic results. Instances are indicated in \textit{italics} with the prefix \textit{ex:}.

Looking at the structure, the EHSA comprises multiple interacting components, including e.g. the Electro-Hydraulic Servo Valve (EHSV), Mode Switching Valve (MSV), two Electro Valves (EVs), Accumulator, and the Actuator Main Ram \cite{Ritter.2018}. Degradations or leakages in these components serve as fault causes that propagate and ultimately impair precise actuator positioning. The \textit{ex:EHSA} is an instance of \texttt{VDI2206:Module}, representing a functional subsystem within the aircraft in order to control the flight surfaces. All further components, such as the \textit{ex:EHSV}, \textit{ex:MSV}, \textit{ex:EV1}, \textit{ex:EV2}, \textit{ex:Accumulator}, and \textit{ex:ActuatorMainRam}, are instances of the class \texttt{VDI2206:Component}, providing elementary sub-functions within the \textit{ex:EHSA}. 

Two operating points, \textit{ex:ActiveMode} (full hydraulic actuation) and \textit{ex:Damping} (bypassing the \textit{ex:EHSV} to restrict movement), are represented in the knowledge graph as instances of the class \texttt{VDI3682:Process}. Process modeling according to VDI 3682 can be carried out using a web-based tool introduced in \cite{Nabizada.2020} and then converted into a graph using a mapping method available on \textit{GitHub}\footnote{\url{https://github.com/hsu-aut/fpb-owl-mapper}}. For simplicity reasons, only \textit{ex:ActiveMode} is depicted in detail in the following. Each instance of the class \texttt{VDI2206:Component} executes a specific \texttt{VDI3682:ProcessOperator} to transform products, energies or signals, enabling precise actuator motion in \textit{ex:ActiveMode}. The process starts with \textit{ex:HydraulicControl}, performed by the \textit{ex:EHSV}, which converts electrical control signals into a hydraulic pressure differential, regulating fluid flow to the actuator chambers. This controlled pressure enables \textit{ex:LinearMotionExecution} in the \textit{ex:ActuatorMainRam}, translating hydraulic energy into mechanical displacement. For adaptive operation the \textit{ex:MSV} determines whether the actuator operates in \textit{ex:ActiveMode} or \textit{ex:DampingMode}. This function is controlled by \textit{ex:EV1} and \textit{ex:EV2}, ensuring smooth state transitions. To maintain system stability, \textit{ex:PressureStabilization}, handled by the \textit{ex:Accumulator}, compensates for pressure fluctuations and ensures continued actuator function during hydraulic supply failures. The \textit{ex:PositionSensor} continuously monitors actuator movement, providing feedback for closed-loop control. Finally, \textit{ex:FailSafeReset}, executed by the \textit{ex:RD}, ensures the \textit{ex:EHSA} returns to its neutral position when no active control signals are present (e.g. during power failures).

\begin{figure}
    \centering
\includegraphics[width=0.7\linewidth]{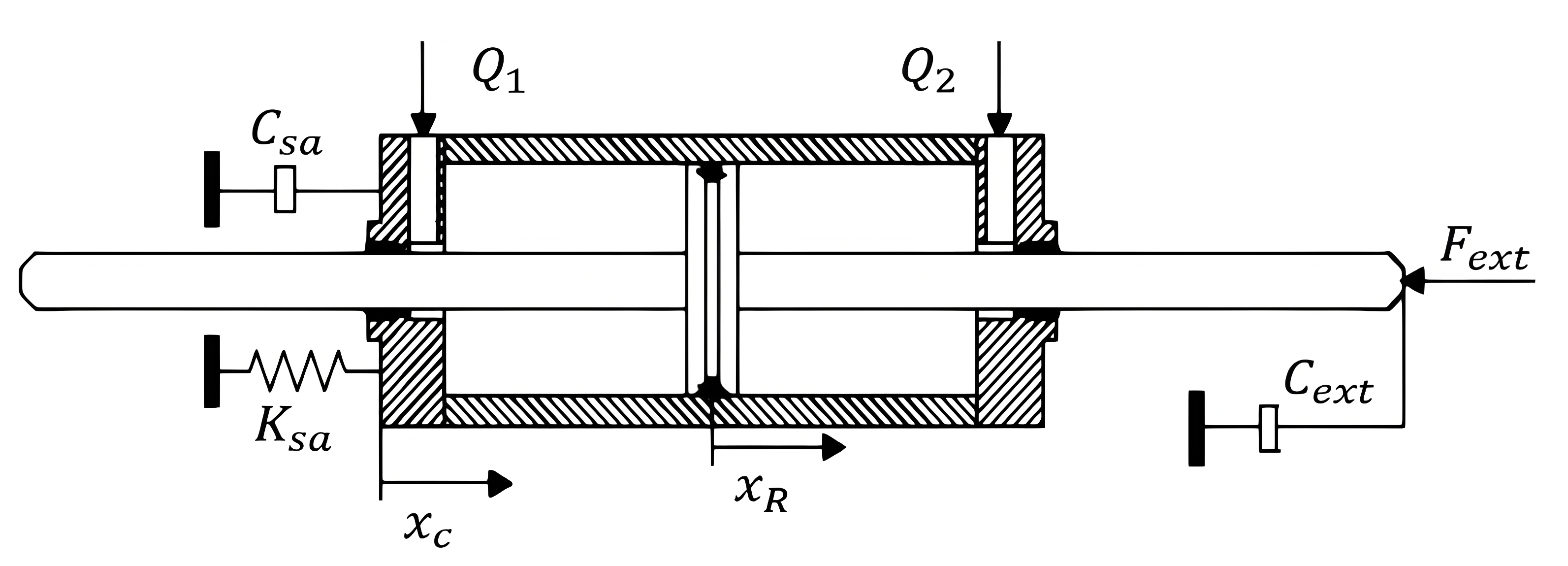}
    \caption{Actuator main ram model, according to \cite{Ritter.2018}}
    \label{fig:Actuator}
\end{figure}

Owing to space constraints, we illustrate the approach using one representative equation in the form of a knowledge graph. The same modeling procedure applies to all other equations and processes. Hence, we instantiate one behavioral model of the \textit{ex:ActuatorMainRam}. Fig.~\ref{fig:Actuator} shows a 2DOF model of the \textit{ex:ActuatorMainRam}, consisting of the actuator ram position $x_R$ and the external sleeve position $x_C$. Their mechanical coupling is represented by a spring ($K_{sa}$) and a damper ($C_{sa}$). An external damper $C_{ext}$ models the interaction at the eye-end of the ram, where the external force $F_{ext}$ is applied. Hydraulic flows $Q_1$ and $Q_2$ control the pressures in the actuator chambers and thereby influence the motion of both masses.
This component plays a central role within the \texttt{VDI3682:ProcessOperator} \textit{ex:LinearMotionExecution}. To mathematically model the transformation of hydraulic energy into mechanical energy, its function must be subdivided into the two subordinate \texttt{VDI3682:ProcessOperator} units \textit{ex:PressureRegularization} and \textit{ex:RamMotionExecution}. The consideration of the process at this granularity level is necessary to consider and simulate leakages within the actuator chamber. All following variables and system parameters (e.g. characteristics of the \textit{ex:ActuatorMainRam}), depicted in Table \ref{tab:EHSA_PressureReg}, were specified semantically using the \texttt{DINEN61360:DataElement}, adding \texttt{DINEN61360:TypeDescription} and \texttt{DINEN61360:InstanceDescription} to these material, energy and signal flows.  \textit{ex:PressureRegularization} receives hydraulic flow inputs $Q_1$ and $Q_2$ from the \textit{ex:EHSV} via the \textit{ex:MSV} and transforms them into pressure rate changes $\frac{\partial p_1}{\partial t}$ and $\frac{\partial p_2}{\partial t}$. Accordingly, the differential pressure, which serve as input for \textit{ex:RamMotionExecution}, drives the motion of the \textit{ex:ActuatorMainRam} at position $x_R$ and the external sleeve at position $x_C$, whose displacement governs the actuator’s position. $\frac{\partial p_1}{\partial t}$ and $\frac{\partial p_2}{\partial t}$ within the actuator chambers evolve dynamically as functions of fluid flow rates $Q_1$ and $Q_2$, hydraulic properties $\beta$, and geometric characteristics $A, V_0$. This relation is captured by (1) and (2) (see Table~\ref{tab:EHSA_PressureReg}). These equations define the forces acting on the ram and thereby influence its mechanical motion, which follows the dynamic equations detailed in \citet{Ritter.2018}.

The relative displacement between the ram and sleeve, given by $x_R$ and $x_C$, determines the actuator's output motion and serves as feedback in the control loop. Leakage effects, both internal ($Q_{\text{li}}$) and external ($Q_{\text{le1}}, Q_{\text{le2}}$), influence hydraulic efficiency.

\begin{table}[h]
    \centering
    \caption{EHSA actuator chamber pressure regularization, inputs, outputs and behavioral models}
    \label{tab:EHSA_PressureReg}
    \renewcommand{\arraystretch}{1.3}
    \begin{tabular}{ll}
        \toprule
        \textbf{Process Operator} & \texttt{PressureRegularization} \\
        \midrule
        \multicolumn{2}{l}{\textbf{Characteristics of Actuator Chamber}} \\
        $V_0$ & Initial chamber volume at neutral position \\
        $A$ & Piston head area \\
        \midrule 
        \multicolumn{2}{l}{\textbf{Inputs}} \\
        $\beta$ & Oil bulk modulus \\
        $Q_1$ & Volume flow into chamber 1 \\
        $Q_2$ & Volume flow into chamber 2 \\
        $Q_{\text{li}}$ & Internal leakage flow between chambers \\
        $Q_{\text{le1}}$ & External leakage flow from chamber 1 \\
        $Q_{\text{le2}}$ & External leakage flow from chamber 2 \\
        $x_R$ & Position of the main ram \\
        $\dot{x}_R$ & Velocity of the main ram \\
        $\dot{x}_C$ & Velocity of the external sleeve \\
        \midrule
        \multicolumn{2}{l}{\textbf{Outputs}} \\
        $\frac{\partial p_1}{\partial t}$ & Pressure rate of change in chamber 1 \\
        $\frac{\partial p_2}{\partial t}$ & Pressure rate of change in chamber 2 \\
        \midrule
        \multicolumn{2}{l}{\textbf{Continuity Equations:}} \\
        \multicolumn{2}{l}{(1) $\frac{\partial p_1}{\partial t}= \beta \frac{Q_1 - Q_{\text{le1}} - Q_{\text{li}} - (\dot{x}_R -  \dot{x}_C) A}{V_0 + x_R A}$} \\
        \multicolumn{2}{l}{(2)  $\frac{\partial p_2}{\partial t}  = \beta \frac{Q_2 - Q_{\text{le2}} + Q_{\text{li}} + ( \dot{x}_R -  \dot{x}_C) A}{V_0 - x_R A}$} \\
        \bottomrule
    \end{tabular}
\end{table}

\begin{figure*}
    \centering
 \includegraphics[width=0.9\linewidth]{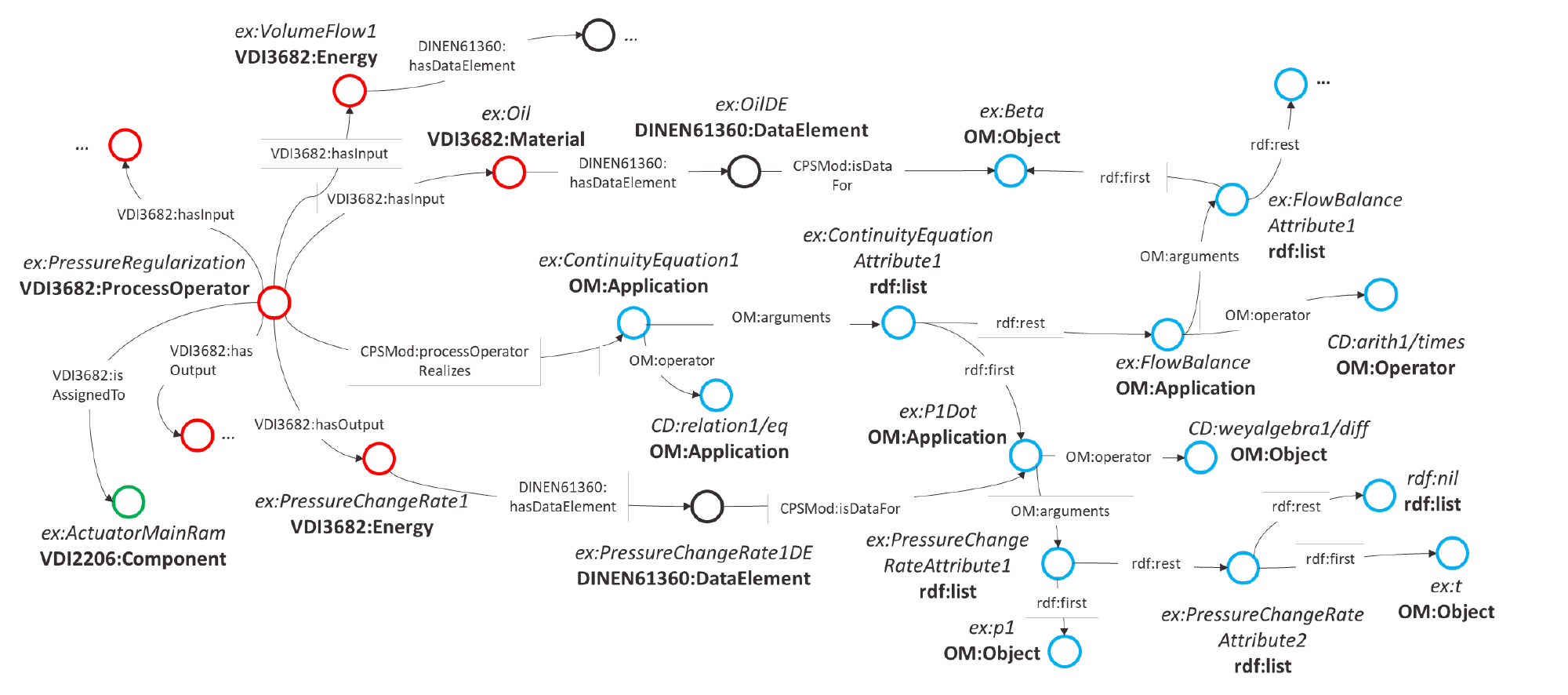}
    \caption{Excerpt from the EHSA knowledge graph} 
    \label{fig:KGEHSA}  
\end{figure*}

After all necessary inputs and outputs as well as the continuity equations have been identified, the next step was to instantiate the mathematical models within the semantic model. For this purpose, the mapping approach, outlined in Sec.~\ref{sec:Modeling}, was followed. To illustrate one instantiated example in RDF serialization, we use the continuity equation (1). Here, an OpenMath XML file that defines the equation in accordance with the standard is used as input to generate an RDF serialization. To simplify this OpenMath XML generation for engineers, mathematical expressions could alternatively be entered using Matlab-style syntax and automatically transformed into OpenMath XML by parsers, which can be implemented with existing libraries. The mapping used here (see Algorithm \ref{alg:OM2RDF}) was implemented in Python using the \texttt{rdflib} library. In this process, the individual sections of the equation are represented using \texttt{OM:Application}, \texttt{OM:Object}, and \texttt{OM:operator} and are linked to \textit{ex:LinearMotionExecution} via the \texttt{VDI2206:MathematicalModel} class. All variables and parameters used in the equation are semantically enriched by linking them to the corresponding \texttt{DINEN61360:DataElement} using the object property \texttt{CPSMod:isDataFor}. An excerpt from the knowledge graph depicting the connected information is shown in Fig. \ref{fig:KGEHSA}. The nodes are colored in accordance to the already introduced semantic model (see Fig. \ref{fig:SemanticModel}). The same systematic instantiation approach is applied to all \texttt{VDI3682:ProcessOperator} within the EHSA described above in order to build the process model. Accordingly, the information stored in this knowledge graph could subsequently be queried to create models in \textit{Matlab/Simulink}, solve the equations, and perform simulations.

%% file: Text/06_Discussion.tex
\section{Discussion} \label{sec:discussion}

The exemplary application shows that the semantic model addresses RQ1 by formalizing complex dynamics, reducing discontinuities, and ensuring interoperability across tools. This is achieved by leveraging OpenMath to integrate ODEs and PDEs into knowledge graphs. Furthermore, by incorporating standards such as VDI 3682 and VDI 2206, these equations are semantically linked to CPS components and functions at various process granularities, providing a composable representation of CPS behavior. Additionally, the proposed method address RQ2 by efficiently instantiating the semantic model with behavior information, reducing manual effort, and enabling systematic integration, querying, and reuse throughout the CPS lifecycle.

Previously, capturing CPS dynamics in knowledge graphs was limited to representing discrete system behavior. The ability to now integrate discrete, continuous, and hybrid behavior brings the approach closer to realistic CPS applications, where complex dynamics must be accurately captured. Furthermore, the integration of time-continuous dynamics into the knowledge graph provides a crucial foundation for interoperability between different engineering tools, enabling e.g. the effective use of Digital Twins.

However, the proposed contribution also has its limitations. It does not support direct computation or numerical simulation, as this is not the intended purpose of ontologies. Instead, the knowledge graph serves purely as an information transfer mechanism, addressing system discontinuities and ensuring interoperability. Computation and numerical solution methods remain external tasks that can be performed using application-specific simulation tools with suited solvers once the information has been retrieved from the knowledge graph. Additionally, the used content dictionary \texttt{weylalgebra1} is still marked as \textit{experimental}, indicating that further standardization efforts are needed to ensure long-term interoperability.

%% file: Text/07_Summary_and_Outlook.tex
\section{Summary and Future Work} \label{sec:conclusion}

CPS exhibit complex dynamic behavior, requiring both time-discrete and time-continuous modeling approaches for accurate representation. To complement the many existing solutions for representing discrete behavior semantically, this paper introduced an approach to integrate differential equations into knowledge graphs.
The two key contributions of this work are as follows:

First, a modular, standards-based semantic model was presented that enables the representation of differential equations using OpenMath. Moreover, the model enhances interoperability and ensures that behavioral models can be efficiently reused and seamlessly integrated with additional information across all lifecycle phases.
Second, a recursive mapping approach for translating OpenMath XML into RDF was introduced to enable the efficient generation of knowledge graphs from mathematical expressions.

Future research will extend the semantic model to represent hybrid CPS behaviors by combining automata-based techniques with differential equations, thus accurately capturing state transitions and event-driven dynamics. Additionally, efforts will be made to automate the model instantiation process by leveraging Large Language Models (LLMs) to extract structured representations from engineering data and system specifications, thereby reducing manual effort. A key focus will also be on exploring data-driven approaches that allow the dynamic adjustment of behavioral models based on sensor data. To strengthen the link between semantic modeling and simulation, further integration with established simulation tools will be explored, supporting the automatic generation of executable models from the knowledge graph.

%% file: bibliography/References.bbl
% Generated by IEEEtranN.bst, version: 1.14 (2015/08/26)